\documentclass[10pt,twocolumn,letterpaper]{article}

\usepackage{iccv}
\usepackage{times}
\usepackage{epsfig}
\usepackage{graphicx}
\usepackage{amsmath}
\usepackage{amssymb}

\usepackage{enumitem}

\usepackage{soul}

\newcommand{\missing}[1]{}

\usepackage{amssymb} %maths
\usepackage{amsmath} %maths
\usepackage{booktabs}
\usepackage{multirow}
\usepackage{bbm}
\usepackage{bm}
\usepackage{mathtools}
\usepackage{lipsum}  
\usepackage[numbers,sort&compress]{natbib}

\newcommand{\nsparagraph}[1]{\noindent \textbf{#1}}
\newcommand{\lsparagraph}[1]{\vspace{0.2cm} \noindent \textbf{#1}}

\DeclareMathOperator*{\argmin}{arg\,min} % Jan Hlavacek
\usepackage{colortbl}
\usepackage{pifont}% http://ctan.org/pkg/pifont
\newcommand{\cmark}{{\ding{51}}}
\newcommand{\xmark}{}

\usepackage{xcolor}
\definecolor{cb-black}      {RGB}{  0,   0,   0}
\definecolor{cb-blue-green} {RGB}{  0,  073,  073}
\definecolor{cb-green-sea}  {RGB}{  0, 146, 146}
\definecolor{cb-rose}       {RGB}{255, 109, 182}
\definecolor{cb-salmon-pink}{RGB}{255, 182, 119}
\definecolor{cb-purple}     {RGB}{ 73,   0, 146}
\definecolor{cb-blue}       {RGB}{ 0, 109, 219}
\definecolor{cb-lilac}      {RGB}{182, 109, 255}
\definecolor{cb-blue-sky}   {RGB}{109, 182, 255}
\definecolor{cb-blue-light} {RGB}{182, 219, 255}
\definecolor{cb-burgundy}   {RGB}{146,   0,   0}
\definecolor{cb-brown}      {RGB}{146,  73,   0}
\definecolor{cb-clay}       {RGB}{219, 209,   0}
\definecolor{cb-green-lime} {RGB}{ 36, 255,  36}
\definecolor{cb-yellow}     {RGB}{255, 255, 109}

\definecolor{maize}{RGB}{255, 180, 0}

\usepackage[pagebackref=true,breaklinks=true,letterpaper=true,colorlinks,bookmarks=false]{hyperref}

\iccvfinalcopy % *** Uncomment this line for the final submission

\ificcvfinal\fi

\begin{document}

%%%%%%%%% TITLE
\title{Bootstrap Your Own Correspondences}

\author{{Mohamed El Banani \qquad Justin Johnson} \\
University of Michigan\\
{\tt\small \{mbanani, justincj\}@umich.edu}
}

\maketitle
\ificcvfinal\thispagestyle{empty}\fi

% stops floats from showing up in the left column
\global\csname @topnum\endcsname 0
\global\csname @botnum\endcsname 0

%%%%%%%%% ABSTRACT
\begin{abstract}
Geometric feature extraction is a crucial component of point cloud registration pipelines.
Recent work has demonstrated how supervised learning can be leveraged to learn better and more compact 3D features.
However, those approaches' reliance on ground-truth annotation limits their scalability.
We propose BYOC: a self-supervised approach that learns visual and geometric features from RGB-D video without relying on ground-truth pose or correspondence. 
Our key observation is that randomly-initialized CNNs readily provide us with good correspondences; allowing us to bootstrap the learning of both visual and geometric features. 
Our approach combines classic ideas from point cloud registration with more recent representation learning approaches. 
We evaluate our approach on indoor scene datasets and find that our method outperforms traditional and learned descriptors, while being competitive with current state-of-the-art supervised approaches.

\end{abstract}

\begin{figure}[t]
\begin{center}
\vspace{0.3cm}
  \includegraphics[width=\linewidth]{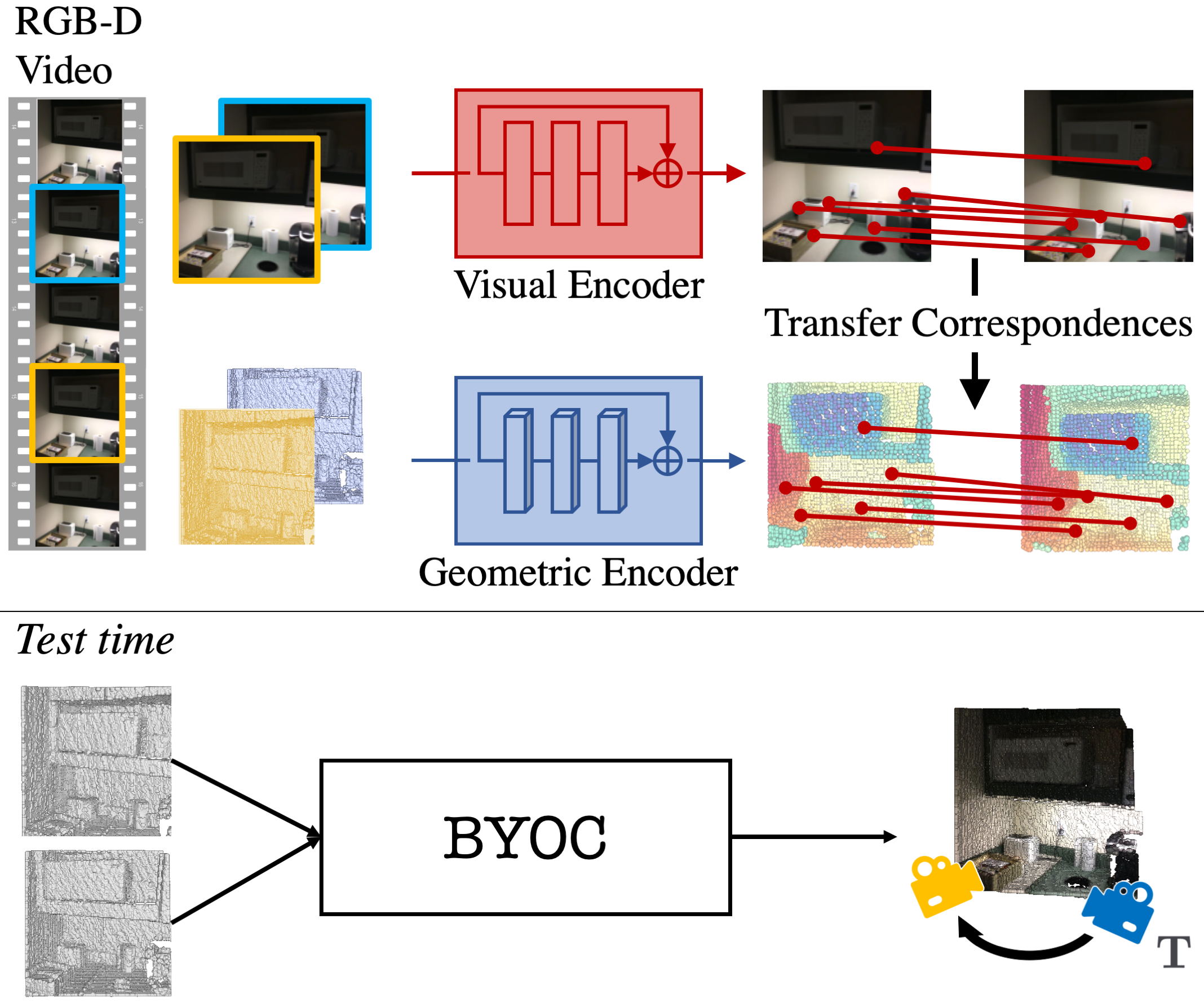}
\end{center}
    \caption{BYOC uses estimated visual correspondence to train a visual and geometric encoder on RGB-D video frames. At test time, it is able to successfully register raw point clouds.}
\label{fig:teaser}
\end{figure}

\section{Introduction}
\label{sec:intro}
One’s ability to align two views of the same scene is closely intertwined with their ability to identify corresponding points between the two  views.
The duality between correspondence estimation and point cloud registration has long been recognized and serves as the basis for many approaches in both problems~\missing{[X]}. 
Given an accurate registration of a scene, one can easily extract correspondences between the two views.
Conversely, given point correspondences, one can easily register two views of a scene.
\textit{Can we leverage this cycle to jointly learn both correspondence estimation and point cloud registration from scratch?}

At the core of this cycle is the ability to generate good feature descriptors for points in the scene.
The prevailing approach to 3D feature learning relies on preregistered scenes to sample ground-truth correspondences for the supervised training of a feature encoder. 
This is done by sampling positive and negative feature pairs and applying triplet~\cite{choy2019fully,khoury2017cgf,Li2020e2e3ddescriptors,yew20183dfeatnet} or contrastive~\cite{bai2020d3feat,choy2019fully,xie2020pointcontrast} losses.
While very successful, these approaches require us to have already registered the raw depth or RGB-D scans to generate the training data. 
This limits this approach to data that can be successfully registered with automated approaches like COLMAP~\cite{schonberger2016structure}.
Ideally, we would leverage the success of supervised approaches without relying on ground-truth correspondence labels. 

To this end, we propose \textbf{B}ootstrap \textbf{Y}our \textbf{O}wn \textbf{C}orrespondences (BYOC): a self-supervised end-to-end approach that learns point cloud registration by leveraging pseudo-correspondence labels. 
Our approach extracts pseudo-correspondence labels from the output of a randomly initialized feature encoder (\S~\ref{sec:approach_pcreg}). 
We use the sampled correspondences to register the point clouds and apply losses based on the quality of the registration to train the feature encoders.
This allows us to slowly bootstrap\footnote{We use \textit{bootstrap} in its idiomatic rather than its statistical sense.} the learning process and learn from RGB-D scans without relying on any pose or correspondence supervision.

This approach works well for registering RGB-D frames, but does worse for aligning point clouds. 
This is primarily due to the fact that randomly initialized 2D CNNs produce more distinctive features than current point cloud encoders, as shown in Figure~\ref{fig:intial_corr}.
We leverage this observation and propose bootstrapping the geometric features using visual correspondences. 
We do this by using the estimated visual correspondences, as opposed to ground-truth labels~\cite{bai2020d3feat,choy2019fully,khoury2017cgf,Li2020e2e3ddescriptors,yew20183dfeatnet,xie2020pointcontrast} to sample positive pairs, and apply a feature metric learning.
We adapt a recently proposed self-supervised method, SimSiam~\cite{chen2020exploring}, for 3D representation learning (\S~\ref{sec:approach_v2g}). 
This addition results in improved performance while being significantly simpler (no negative samples or momentum encoders) than prior contrastive learning formulations applied to point clouds. 

Our work draws inspiration from two sources: iterative closest point algorithm (ICP)~\cite{besl1992method,chen1992object,zhang1994iterative} and self-supervised learning on pseudo-labels~\cite{caron2018deep,grill2020byol,lee2013pseudo}. 
While seemingly different, the same intuition lies at the core of both lines of work.
ICP is a registration algorithm that assumes that the closest points between two point clouds correspond to each other. 
Through iterative refinement and resampling, it can register roughly aligned point clouds. 
Meanwhile, self-supervised learning with pseudo-labels learns to predict pseudo-labels in the form of current top prediction~\cite{lee2013pseudo}, feature clusters~\cite{caron2018deep}, or even a previous prediction~\cite{grill2020byol}. 
Through redefining the labels over time, the model can learn good representations. 
Both rely on the observation that pseudo-labels in a well-structured space (\ie, similar objects already lie close to each other) can provide a valuable learning signal. 
This is particularly relevant for learning due to the finding that CNNs, \textit{even} when randomly initialized, still serve as good feature extractors~\cite{rosenfeld2019intriguing,ulyanov2018deep}. 

We evaluate our approach on two indoor scene datasets: ScanNet~\cite{dai2017scannet} and 3D Match~\cite{zeng20163dmatch}. 
Despite the simplicity of our approach, it outperforms hand-crafted features as well as several supervised baselines, while being competitive with current state-of-the-art approaches.  

In summary, we propose a self-supervised approach that uses sampled correspondences from initially random feature encoders to learn point-wise features for point cloud registration (\S~\ref{sec:approach_pcreg}). We also demonstrate how visual correspondences could be used to further improve geometric feature learning (\S~\ref{sec:approach_v2g}).
We demonstrate the efficacy of this approach on point cloud registration (\S~\ref{sec:exp_pcreg}) and correspondence estimation (\S~\ref{sec:exp_corr}).

\section{Related Work}
\label{sec:related}
\paragraph{3D Feature Descriptors.}
Early work on feature point extraction can be traced back to using corners for stereo matching~\cite{moravec1981rover}. 
The core intuition of extracting features based on histograms of gradients was extended to 3D features~\cite{johnson1997spin, johnson1999using, rusu2009fast, salti2014shot, tombari2010usc}.
More recently, there has been a nascent body of work focused on leveraging supervised learning for learning 3D features~\cite{bai2020d3feat,choy2019fully,deng2018ppffoldnet,deng2018ppfnet,deng2019directreg,gojcic2019perfect,khoury2017cgf,Li2020e2e3ddescriptors,wang2019dcp,yew20183dfeatnet,zhao20193dpointcaps}.  
The common approach is to sample both positive and negative pairs between two frames and then use them in a triplet~\cite{choy2019fully,khoury2017cgf,Li2020e2e3ddescriptors,yew20183dfeatnet}, contrastive~\cite{bai2020d3feat,choy2019fully,xie2020pointcontrast}, or an N-tuple~\cite{deng2018ppfnet} loss.
Other approaches propose unsupervised learning approaches on reconstructed scenes~\cite{deng2018ppffoldnet,xie2020pointcontrast,zhao20193dpointcaps}. 
While those approaches do not explicitly use ground-truth pose, they rely on reconstructed scenes which are generated using ground-truth pose. 
Unlike prior approaches, our approach operates directly on depth or RGB-D scans without relying on ground-truth pose or correspondence and focuses on point cloud registration as an end task.

\lsparagraph{Point Cloud Registration.} 
Early work on point cloud registration assumed perfect correspondence between the point clouds~\cite{arun1987least,longuet1981computer}.
This assumption was later relaxed by ICP by assuming the closest point is the correspondence~\cite{besl1992method,chen1992object,zhang1994iterative}.
While this assumption holds for several applications (\eg, registering scans from a high frame-rate scanner or fine-tuning alignment), it is challenged by large transformations and partially overlapping point clouds. 
Later on, feature-based approaches were proposed that compute features to establish correspondence, and use robust estimators such as RANSAC to handle noise and outliers~\cite{torr1997development, zhang1995robust}. For a review, see ~\cite{pomerleau2015review}. 
More recent approaches incorporate learning into the registration process~\cite{brachmann2017dsac, brachmann2019neural,choy2020deep,elbanani2021unsupervisedrr,huang2020featureregistration,gojcic2020learning,ranftl2018deepfundamental,yew2020RPMNet}.
Finally, there has been a line of work that proposes self-supervised approaches for registering objects~\cite{aoki2019pointnetlk,hertz2020pointgmm,huang2020featureregistration,wang2019dcp,wang2019prnet,yew2020RPMNet,yuan2020deepgmr} or reconstructed scenes~\cite{khoury2017cgf,deng2018ppffoldnet,zhao20193dpointcaps}. 
We are inspired by this line of work, but differ from it in two key ways: our approach operates directly on raw depth or RGB-D scans and is unsupervised.

\lsparagraph{Self-supervised learning.}
Self-supervised learning refers to approaches that apply supervised learning to tasks where the data itself serves as the supervision. This idea has been very popular for 2D representation learning with the goal of learning representations that generalize to downstream tasks~\cite{chen2020exploring,desai2020virtex,doersch2015unsupervised,gidaris2018unsupervised,grill2020byol,goyal2019scaling,tian2019contrastive}. Recently, PointContrast~\cite{xie2020pointcontrast} and DepthContrast~\cite{zhang2021depthcontrast} demonstrated how to extend this formulation to 3D representation learning. We are inspired by this line of work but differ from it in several ways. First, our goal is to use self-supervision to tackle the task itself, not a different downstream task. Second, we differ from prior 3D self-supervision in that we operate on depth scans, not reconstructed scenes like~\cite{xie2020pointcontrast}. Also, we learn point-wise representations, not holistic representations like~\cite{zhang2021depthcontrast}. 
Finally, our work is inspired by the recently proposed SimSiam~\cite{chen2020exploring} which proposes a self-supervised approach that is much simpler than prior approaches. We adapt SimSiam to the point cloud setting by applying it to the sampled feature pairs.

\begin{figure*}[ht]
\begin{center}
  \includegraphics[width=0.8\linewidth]{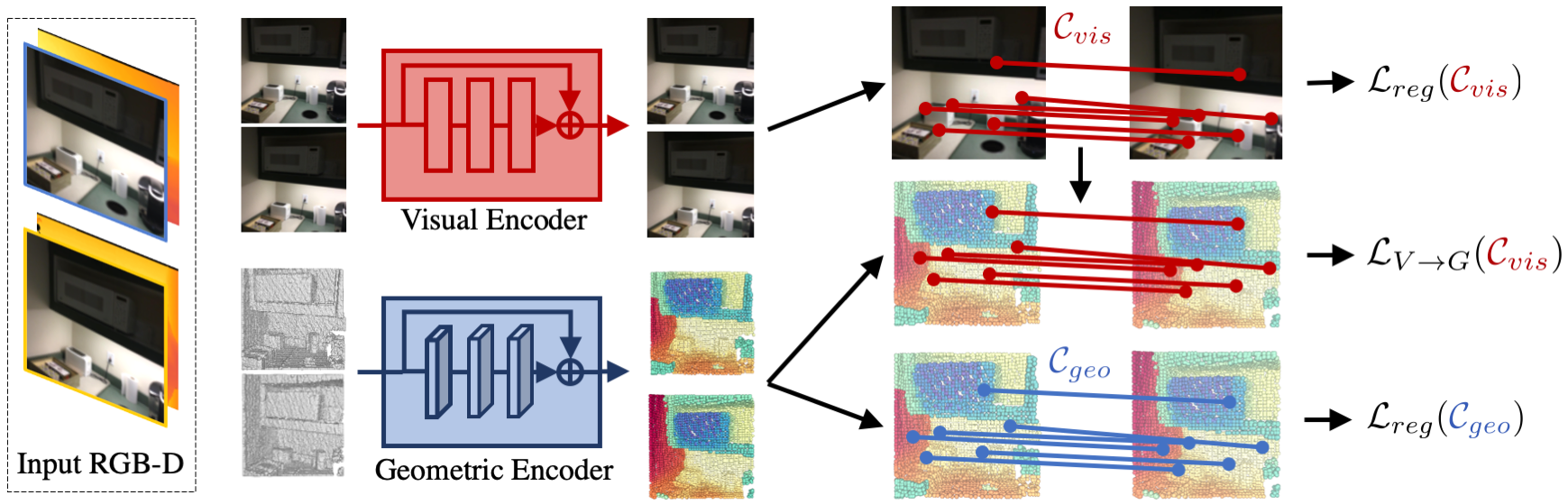}
\end{center}
    \caption{\textbf{BYOC}. 
    Our model takes as input two RGB-D images of a scene. First, we extract visual features from the images and geometric features from the point clouds. This results in two point clouds where each point has a 3D location, visual feature, and geometric feature. We then extract correspondences from the visual and geometric features. Those correspondences are used to estimate a transformation and compute a registration loss. We also apply a feature similarity loss on geometric features sampled using the visual correspondences.
    }
\label{fig:model}
\end{figure*}

\section{Approach}
\label{sec:approach}
The goal of this work is to learn geometric point cloud registration from RGB-D video without relying on pose or correspondence supervision. 
Our approach, shown in Fig.~\ref{fig:model}, has three major components: visual registration, geometric registration, and correspondence transfer. 
The first two components are based on the traditional registration pipeline of feature extraction, correspondence estimation, and geometric fitting. 
The only difference between them is whether the features are extracted using a visual encoder from the image or a geometric encoder from the point cloud. 
The third component is based on SimSiam~\cite{chen2020exploring}, where we apply a loss on pairs of geometric features that are sampled using visual correspondences. 
Our key insight is that randomly initialized CNNs produce features that allow for decent correspondence estimation. 
This allows us to bootstrap the learning of both visual and geometric encoders by using estimated correspondences with  supervised registration and feature similarity losses.

\subsection{Point Cloud Registration} 
\label{sec:approach_pcreg}

Given two point clouds, $\mathcal{P}_0$ and $\mathcal{P}_1$, point cloud registration is the task of finding the transformation $\mathbf{T} \in \text{SE}(3)$ that aligns them. 
Registration approaches commonly consist of three stages: feature extraction, correspondence estimation, and geometric fitting. 
In our approach, we perform two registration steps using image or point cloud features. Correspondence estimation and geometric fitting are the same for both steps. 
Below we discuss each of the steps in detail.

\lsparagraph{Geometric Feature Extraction. }
Our first encoder takes point clouds as input. This encoder allows us to extract features based on the geometry of the input scene. 
We first generate a point cloud for each view using the input depth and known camera intrinsic matrix.
We then encode each point cloud using a sparse 3D convolutional network~\cite{choy2019minkowski,Graham_2018_CVPR}. 
We use this network due to its success as a back-end for supervised registration approaches~\cite{choy2019fully,choy2020deep,gojcic2020learning} and 3D representation learning~\cite{xie2020pointcontrast,zhang2021depthcontrast}. 
This network applies sparse convolution to a voxelized point cloud, which allows it to extract features that capture local geometry while maintaining a quick run-time. 
Similar to prior work~\cite{choy2019fully,xie2020pointcontrast,zhang2021depthcontrast}, we find that a voxel size of 2.5 cm works well for indoor scenes. 
This step maps our input RGB-D image, $I_0, I_1 \in  \mathbb{R}^{4 \times H \times W}$ to $\mathcal{P}_0, \mathcal{P}_1 \in \mathbb{R}^{N \times (3 + F)}$ where each point cloud has N points, and each point, $p$, is represented by a 3D coordinate, $\mathbf{x}_p$, and a $F$-dimensional geometric feature vector, $\mathbf{g}_p$.\footnote{Voxelization will result in point clouds of varying dimension. We use heterogeneous batching to handle this in our implementation, but assume that point clouds have the same size in our discussion for clarity.} In our experiments, we use a feature dimension of 32.

\lsparagraph{Visual Feature Extraction. }
Our second encoder takes images as input and generates an output feature map of the same size. 
Maintaining the image's spatial resolution results in a feature vector extracted for each pixel. 
We use a ResNet encoder with two residual blocks as our image encoder and map each pixel to a feature vector of size 32. 
We use the projected 3D coordinates of the voxelized point cloud from the geometric encoder to index into the 2D feature map. 
This allows us to generate a point cloud for each input RGB-D image whose points $p \in \mathcal{P}$ have both a visual feature vector, $\mathbf{v}_p$, and a geometric feature vector, $\mathbf{g}_p$. 
We use this property for transferring correspondences between the different feature modalities in \S~\ref{sec:approach_v2g}. 
We only use the visual encoder during training to bootstrap the geometric feature learning. At test time, we register point clouds without access to image data.

\lsparagraph{Correspondence estimation. }
We estimate the correspondences between the two input views for each feature modality to output two sets of correspondences: $\mathcal{C}_{vis}$ and $\mathcal{C}_{geo}$. 
We first generate a list of correspondences by finding the nearest neighbor to each point in the feature space. 
Since we have two point clouds of N points, we end up with a correspondence list of length 2N candidate correspondences.

The candidate correspondences will likely contain a lot of false positives due to poor matching, repetitive features, and occluded or non-overlapping portions of the image. 
The common approach is to filter the correspondences based on some criteria of uniqueness or correctness. 
Recent approaches propose learning networks that estimate a weight for each correspondence~\cite{choy2020deep,gojcic2020learning,ranftl2018deepfundamental}.
In this work, we leverage the method proposed by~\cite{elbanani2021unsupervisedrr} of using a weight based on Lowe's ratio~\cite{lowe2004distinctive}. 
Given two point clouds, $\mathcal{P}_0$ and $\mathcal{P}_1$, we find the correspondences of point $p \in \mathcal{P}_0$ by finding the two nearest neighbors $q_p$ and $q_{p, nn_2}$ to $p$ in $\mathcal{P}_1$ in feature space. 
We can calculate the Lowe's ratio weight as follows: 
\begin{equation}
    w_{p, q_p} =  1 - \frac{D(\mathbf{f}_p, \mathbf{f}_{q_p})}{D(\mathbf{f}_p, \mathbf{f}_{q_{p, nn_2}})}
\end{equation}
where $D$ is cosine distance, and $\mathbf{f}_p$ is either the visual or the geometric feature descriptor depending on which correspondence set is being calculated.
It is worth noting that this formulation is similar to the triplet loss often used in contrastive learning, where $q_p$ is the positive sample and $q_{p, nn_2}$ is the hardest negative sample. 
We use the resulting weights to rank the correspondences and only include the top $k$ correspondences. We use $k=400$ in our experiments. 
Each element of our correspondence set $\mathcal{C}$ consists of the two corresponding points and their weight $(p, q, w_{p, q})$.

\lsparagraph{Geometric Fitting. }
For each set of correspondences, we estimate the transformation, $\mathbf{T}^{*} \in \text{SE(3)}$ that would minimize the mean-squared error between the aligned correspondences:
\begin{equation}
{E}(\mathcal{C}, \mathbf{T}) = \frac{1}{|\mathcal{C}|} \sum_{(p, q_p, w) \in \mathcal{C}} \frac{w}{\sum_{\mathcal{C}} w}||\mathbf{x}_{q_p} - \mathbf{T}(\mathbf{x}_p)||
\end{equation}
% \begin{equation}
% \mathbf{T}^{*} = \argmin_{\mathbf{T} \in \text{SE}(3)}{E(\mathcal{C}, \mathbf{T})}
% \end{equation}
Choy~\etal~\cite{choy2020deep} show this problem can be reformulated as a weighted Procrustes algorithm~\cite{gower1975generalized,kabsch1976solution}, allowing for weights to be integrated into the operation to improve the optimization process while maintaining differentiability with respect to the weights. 
We adopt this formulation due to its relative simplicity and ease of incorporation within an end-to-end trainable system. 
Despite having a filtered correspondence list, the correspondence set might still include some outliers that would result in an incorrect geometric fitting.
We adopt the randomized optimization used in~\cite{elbanani2021unsupervisedrr}, and similarly find that we get the best performance by only using it at test time.

\begin{figure}[t]
\begin{center}
  \includegraphics[width=0.85\linewidth]{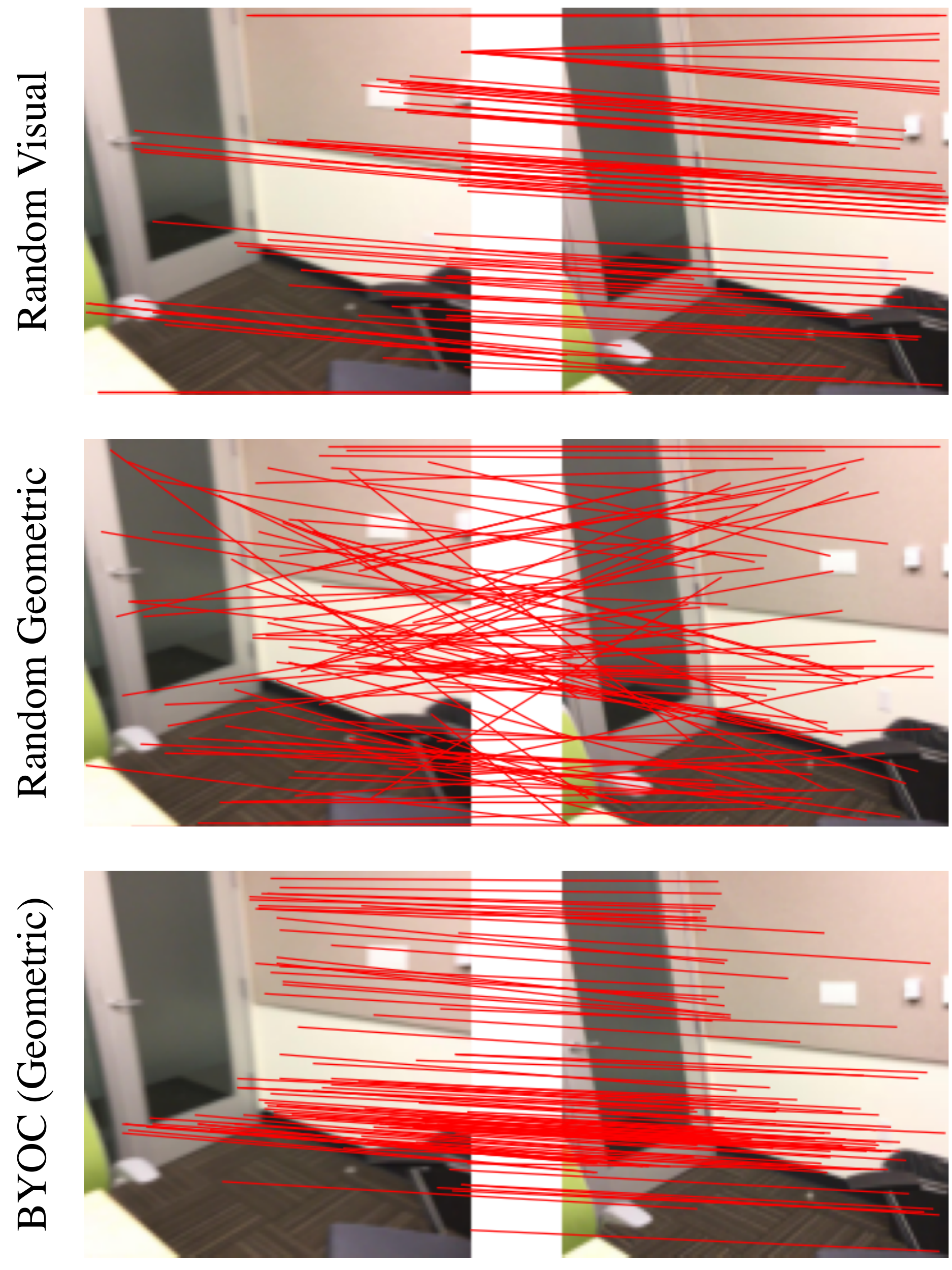}
\end{center}
    \caption{\textbf{Randomly-initialized CNN features are good feature descriptors.} 
    We can estimate good correspondences from random visual features, but not random geometric features. 
    We leverage this observation to bootstrap the learning of geometric features using visual correspondences. 
    Our learned approach learns to estimate accurate correspondence from geometric features. 
 }
\label{fig:intial_corr}
\end{figure}

\lsparagraph{Registration Loss. }
Our registration loss is defined with respect to our correspondence set and the estimated transformation as follows: 
\begin{equation}
\mathcal{L}_{reg}(\mathcal{C}) = \argmin_{\mathbf{T} \in \text{SE}(3)}{E(\mathcal{C}, \mathbf{T})}
\end{equation}
There are a few interesting things about this loss. First, it is worth noting that the gradients are back-propagated to the feature encoder through the weights, $w$, \textit{and} the transformation, $\mathbf{T}$. Hence, the loss can be formulated without using the weights. We find that using the weight improved the performance of visual registration while deteriorating the performance of geometric registration. Therefore, in our model, we only apply the weighting to the visual registration branch while removing it from the geometric branch. 

Second, the loss operates as a weighted sum over the residuals, specifically, the loss is minimized if the correspondence with the lowest residual error has the highest weight. Since the weights are L1 normalized, the relative weighing of the correspondences matters. Removing the normalization results in an obvious degeneracy since the loss can be minimized by driving the weights to 0, which can be achieved by mode collapse. Finally, the weighted loss closely resembles a triplet loss since we sample both positive (first nearest neighbor) and hardest negative (second nearest neighbor) samples. However, unlike the commonly used margin triplet loss, this formulation does not require defining a margin as it operates on the ratio of distances rather than their absolute scale.

\subsection{Visual $\to$ Geometric}
\label{sec:approach_v2g}

The approach outlined in \S~\ref{sec:approach_pcreg} works well with visual features, however it is less effective with geometric features. 
The reason for this becomes apparent once we consider the registration performance using features from randomly initialized encoders. 
As shown in Figure~\ref{fig:intial_corr}, we observe that the features extracted using randomly initialized visual encoder provide some distinctive output, while a random geometric encoder's outputs are more random.

Ideally, we would leverage the good visual correspondences to further bootstrap the learning of the geometric features. 
We observe that geometric feature learning approaches typically define metric learning losses using sampled correspondences~\cite{bai2020d3feat,choy2019fully,gojcic2019perfect,Li2020e2e3ddescriptors,yew20183dfeatnet}.
We adapt this approach to the unsupervised setting by sampling feature pairs using visual correspondences.
This is fairly simple in our approach since each point has both a visual feature and a geometric feature, so transferring correspondences becomes as simple as indexing into another tensor. 
Since the correspondences act as an indexing mechanism, the loss is only back-propagated to the geometric encoder.

Current 3D feature learning approaches rely on both positive and negative pairs to define triplet~\cite{choy2019fully,khoury2017cgf,Li2020e2e3ddescriptors,yew20183dfeatnet} or  contrastive~\cite{bai2020d3feat,choy2019fully,xie2020pointcontrast} losses.
However, as noted in the literature, those losses can be difficult to apply due to their susceptibility to mode collapse and sensitive to hyperparameter choices and negative sampling strategy~\cite{choy2019fully,xie2020pointcontrast,zhang2021depthcontrast}. 
Those issues are amplified in our setting since we rely on estimated, not ground-truth, correspondences. 
Instead of the typical contrastive setup, we adapt the recently proposed SimSiam~\cite{chen2020exploring} to the point cloud setting. 
SimSiam allows us to train our model without requiring negative sampling or having any hyperparameters, while being less susceptible to mode collapse than contrastive losses~\cite{chen2020exploring}.

We adapt SimSiam by applying it to the geometric features of corresponding points, instead of features for different augmentations of the same image. 
Given a correspondence $(p, q) \in \mathcal{C}_{vis}$, we first project the features using a two-layer MLP projection head and apply a stop-gradient operator on the features:
\begin{equation}
\mathbf{z}_p = \mathtt{stopgradient}(project(\mathbf{g}_p)).
\end{equation}
We then compute the loss based on the cosine distance between each geometric feature and the projection of its correspondence:
\begin{equation}
\mathcal{L}_{V\to G}(\mathcal{C}_{vis}) = \frac{1}{|\mathcal{C}_{vis}|} \hspace{-0.5mm} \sum_{(p, q) \in \mathcal{C}_{vis}} \hspace{-4mm} D(\mathbf{g}_p, \mathbf{z}_q) + D(\mathbf{g}_q, \mathbf{z}_p) 
\end{equation}
where $D$ is the cosine distance function and $\mathcal{C}_{vis}$ is the set of visual correspondences. 
In our experiments, we observe that adding SimSiam improved the performance without requiring any additional fine-tuning or model changes.

\begin{table*}[t]
\centering
\resizebox{\textwidth}{!}{

\begin{tabular}{l cc  ccccc ccccc ccccc}
\toprule
&  & & \multicolumn{5}{c}{Rotation} & \multicolumn{5}{c}{Translation} & \multicolumn{5}{c}{Chamfer}\\

& &  
& \multicolumn{3}{c}{Accuracy $\uparrow$} & \multicolumn{2}{c}{Error $\downarrow$} 
& \multicolumn{3}{c}{Accuracy $\uparrow$} & \multicolumn{2}{c}{Error $\downarrow$} 
& \multicolumn{3}{c}{Accuracy $\uparrow$} & \multicolumn{2}{c}{Error $\downarrow$} \\ 

\cmidrule(lr){ 4- 6}   \cmidrule(lr){ 7-8}
\cmidrule(lr){ 9-11}   \cmidrule(lr){12-13}
\cmidrule(lr){14-16}   \cmidrule(lr){17-18}
  
&  Train Set & Pose Sup.  
& $5^\circ$ & $10^\circ$ & $45^\circ$ & Mean & Med. 
& 5 & 10 & 25 & Mean & Med. 
& 1 & 5 & 10 & Mean & Med. 
\\
\midrule

ICP (Point-to-Point) & - & \xmark & 
31.7 & 55.6 & 99.6 & 10.4 &  8.8 &  7.5 & 19.4 & 74.6 & 22.4 & 20.0 &  8.4 & 24.7 & 40.5 & 32.9 & 14.1 
\\

ICP (Point-to-Plane) & - & \xmark & 
54.4 & 68.0 & 98.6 &  8.6 &  3.6 & 30.0 & 36.7 & 70.4 & 23.6 & 18.0 & 31.6 & 43.1 & 53.5 & 229.5 &  8.2
\\

FPFH~\cite{rusu2009fast} + Weighted Procrustes & - & \xmark &
22.2 & 48.2 & 84.9 & 27.8 & 10.4 &  7.4 & 19.6 & 56.3 & 54.1 & 25.3 & 17.5 & 46.8 & 61.2 & 26.5 &  5.8
\\

FPFH~\cite{rusu2009fast} + RANSAC & - & \xmark & 
34.1 & 64.0 & 90.3 & 20.6 &  7.2 &  8.8 & 26.7 & 66.8 & 42.6 & 18.6 & 27.0 & 60.8 & 73.3 & 23.3 &  2.9
\\

\midrule
FCGF~\cite{choy2019fully} + Weighted Procrustes \hspace{0.3cm} & 3D Match & \cmark &
54.1 & 73.3 & 92.2 & 15.3 &  4.3 & 30.8 & 46.2 & 73.0 & 35.0 & 11.6 & 45.6 & 67.4 & 76.4 & 21.5 &  1.4
\\

FCGF~\cite{choy2019fully} + RANSAC & 3D Match & \cmark  &
75.3 & 87.7 & 95.6 &  9.7 &  2.5 & 39.7 & 64.9 & 86.5 & 20.8 &  6.4 & 62.5 & 83.1 & 88.2 & 13.0 &  0.6 
\\

FCGF~\cite{choy2019fully} + DGR~\cite{choy2020deep} & 3D Match & \cmark  &
83.6 & 90.5 & 95.2 &  9.0 &  1.7 & 57.6 & 78.8 & 91.3 & 17.1 &  4.2 & 76.5 & 89.4 & 91.8 & 10.7 &  0.3

\\
FCGF~\cite{choy2019fully} + 3D MV Reg~\cite{gojcic2020learning} & 3D Match & \cmark  &
87.7 & 93.2 & 97.0 &  6.0 &  1.2 & 69.0 & 83.1 & 91.8 & 11.7 &  2.9 & 78.9 & 89.2 & 91.8 & 10.2 &  0.2 
\\

\midrule

BYOC & 3D Match & \xmark  &  
66.5 & 85.2 & 97.8 &  7.4 &  3.3 & 30.7 & 57.6 & 88.9 & 16.0 &  8.2 & 54.1 & 82.8 & 89.5 &  9.5 &  0.9
\\

BYOC-Rot & ScanNet & \xmark & 
71.2 & 87.8 & 97.9 &  6.9 &  2.8 & 38.2 & 63.3 & 89.8 & 15.0 &  6.8 & 61.8 & 84.7 & 90.7 &  9.3 &  0.6
\\

BYOC-Geo & ScanNet & \xmark & 
80.3 & 92.8 & 98.8 &  4.8 &  2.3 & 46.5 & 74.6 & 94.6 & 10.6 &  5.4 & 71.9 & 91.1 & 94.5 &  7.2 &  0.5
\\
    
BYOC + RANSAC & ScanNet & \xmark  &
81.3 & 92.8 & 98.4 &  5.6 &  2.4 & 37.8 & 69.7 & 92.1 & 13.3 &  6.4 & 67.7 & 89.8 & 93.5 &  7.7 &  0.5
\\

BYOC  & ScanNet & \xmark  &
86.5 & 95.2 & 99.1 &  3.8 &  1.7 & 56.4 & 80.6 & 96.3 &  8.7 &  4.3 & 78.1 & 93.9 & 96.4 &  5.6 &  0.3
\\

\bottomrule
\end{tabular}
}
\label{tab:pose_scannet_fake}
\caption{\textbf{Pairwise Registration on ScanNet.}
We outperform existing registration pipelines that use traditional and learned geometric feature descriptors with a RANSAC or Weighted Procrustes estimator. 
Furthermore, we perform on-par with supervised approaches that were trained on 3D Match, demonstrating the utility of unsupervised training in this domain.
\textit{Pose Sup.} indicates pose supervision.
}
\label{tab:pose_scannet}

\end{table*}

\section{Experiments}
\label{sec:experiments}
We evaluate our approach on point cloud registration of indoor scenes.
We train our model on ScanNet, a large dataset of indoor scenes, and evaluate it on ScanNet and the 3D Match registration benchmark. 
Our experiments aim to answer the following questions: 
(1) can we learn accurate point cloud registration from bootstrapped correspondences?;
(2) can we leverage RGB-D video at training time to train better geometric encoders?

\lsparagraph{BYOC variants. }
We consider three variants of our model: BYOC, BYOC-Geo, and BYOC-Rot. 
The first variant of our full model, depicted in Fig.~\ref{fig:model}, is trained using RGB-D pairs, but only uses the geometric encoder at test time to register point clouds. 
The second variant, BYOC-Geo, is trained on depth pairs with only the registration loss on geometric correspondences. This variant applies the bootstrapping idea without further leverage the visual correspondence. 
Our third variant, BYOC-Rot, is the full model trained with additional rotation augmentation. We sample random rotations and apply them to the point cloud before the geometric encoder. This is a common form of augmentation in 3D feature learning~\cite{choy2019fully, xie2020pointcontrast} and is intended to provide the learned feature with some rotational equivariance. 

% -- Experimental Setup
\lsparagraph{Datasets.} 
We evaluate our approach on two datasets of indoor scenes: ScanNet~\cite{dai2017scannet} and 3D Match~\cite{zeng20163dmatch}. 
While both datasets provide RGB-D video annotated with ground-truth camera poses, 3D Match provides an additional geometric registration benchmark that is more challenging due to the larger viewpoint changes. 
ScanNet provides pose annotated RGB-D video for 1513 scenes, while 3D Match's RGB-D video dataset only spans 101 scenes. 
We emphasize that we only use RGB-D video and camera intrinsics for training our model.
We use the official train/valid/test scene split for both datasets, and generate view pairs by sampling image pairs that are 20 frames apart. 
This results in 1594k/12.6k/26k RGB-D pairs for ScanNet and 122k/1.5k/1.5k RGB-D pairs for 3D Match.

\lsparagraph{Training Details. }
We train our model with the Adam~\cite{kingma2015adam} optimizer using a learning rate of $10^{-4}$ and momentum parameters of (0.9, 0.99). 
We train each model for 200K iterations with a batch size of 8. We implement our models in PyTorch~\cite{pytorch3d}, while making extensive use of PyTorch3D~\cite{pytorch3d}, Open3D~\cite{open3d}, and Minkowski Engine~\cite{choy2019minkowski}. 

% Point Cloud Registration
\subsection{Point Cloud Registration}
\label{sec:exp_pcreg}

We first evaluate our approach on point cloud registration on ScanNet and report our results in Table~\ref{tab:pose_scannet}. 
Given two point clouds, we estimate the transformation $\mathbf{T} \in \text{SE}(3)$ that would align the point clouds. 
We emphasize that we discard the visual encoder at test time, and only use the geometric encoder on point cloud input. 

\lsparagraph{Baselines.} While our approach is unsupervised, we are interested in comparing to both classical hand-crafted and supervised learning approaches. 
We first compare our approach against different variants of ICP~\cite{rusinkiewicz2001efficient}.
ICP is an important comparison since it is both an inspiration of this work, as well as a classical algorithm for point cloud registration. 
We also compare it with a RANSAC-based aligner using FPFH~\cite{rusu2009fast} or FCGF~\cite{choy2019fully} 3D feature descriptors. 
FPFH~\cite{rusu2009fast} is a hand-crafted 3D feature descriptor that encodes a histogram of the geometric relationships between each point and its nearest neighbors. 
FPFH is one of the best non-learned 3D feature descriptors and would be representative of the performance of hand-crafted 3D features. 
FCGF~\cite{choy2019fully} is a recently proposed learned 3D feature descriptor that combines sparse 3D convolutional networks with contrastive losses defined on ground-truth correspondences to achieve state-of-the-art performance on several registration benchmarks. 
Finally, we compare with Deep Global Registration~\cite{choy2020deep} and 3D Multiview Registration~\cite{gojcic2019perfect}: two supervised approaches that learn to estimate correspondences on top of FCGF features. 
Those approaches use supervision for both feature learning and correspondence estimation, while our approach is unsupervised for both. It is worth noting that 3D Multi-view Registration~\cite{gojcic2020learning} proposes both a method for pairwise registration and synchronizing multiple views at the same time. We only compare against their pairwise registration module. 

\lsparagraph{Evaluation Metrics. }
We evaluate the pairwise registration by calculating the rotation and translation error between the predicted and ground-truth transformation as follows:   
\begin{equation}
E_{\text{rotation}} = \arccos(\frac{Tr(\mathbf{R}_{pr}\mathbf{R}_{gt}^\top) - 1}{2}), 
\end{equation}
\begin{equation}
E_{\text{translation}} = ||\mathbf{t}_{pr} - \mathbf{t}_{gt}||_2 .
\end{equation}
We report the translation error in centimeters and the rotation errors in degrees. We also report the chamfer distance between the predicted and ground-truth alignments of the scene. 
For each of metric, we report the mean and median errors as well as the accuracy at different thresholds.

\nsparagraph{Results. }
We first note that ICP approaches fail on this task. ICP assumes that the point clouds are prealigned and can be very effective at fine-tuning such alignment by minimizing a chamfer distance. However, our view pairs have a relatively large camera motion with the mean transformation between two frames being 11.4 degrees and 19.4 cm. 
As a result, ICP struggles with the large transformations and partial overlap between the point cloud pairs.
Similarly, FPFH also fails on this task as its output descriptors are not distinctive enough, resulting in many false correspondences which greatly deteriorates the registration performance.

On the other hand, learned approaches show a clear advantage in this domain as they are able to learn features that are well-tuned for the task and data domain. 
Our model is able to outperform FCGF despite FCGF being trained with ground-truth correspondences on an indoor scene dataset.
This is true regardless of whether our model is trained using RGB-D or depth pairs. 
While we find that our model trained on 3D Match performs worse than FCGF, this is expected since 3DMatch is a much smaller dataset making it less suitable for a self-supervised approach. 

Finally, our approach is competitive against approaches that use supervision for both feature learning and correspondence estimation~\cite{choy2020deep,gojcic2020learning}. 
This comparison represents the difference between full supervision on a small dataset vs. self-supervision on a large dataset. 
Our competitive performance demonstrates the promise of self-supervision in this space and our model's ability to learn for a very simple learning signal: consistency between video frames. 

{
\begin{table}[t]
\centering
\resizebox{\linewidth}{!}{

    \begin{tabular}{l | cc cc cc }
    \toprule
    
 \multicolumn{1}{c}{} & \multicolumn{2}{c}{Rotation Error} & \multicolumn{2}{c}{Translation Error} &\multicolumn{2}{c}{Chamfer}  \\
\cmidrule(lr){ 2- 3}   \cmidrule(lr){ 4-5} \cmidrule(lr){6-7}
 \multicolumn{1}{c}{}  & Mean & Median & Mean & Median & Mean & Median  \\
\midrule
Random Visual       &   6.4 &  2.7 & 14.9 &  7.0 & 9.8 &  0.6    \\
Random Geometric    &  21.3 & 13.0 & 46.5 & 28.5 & 26.0 &  8.6  \\
\midrule 
BYOC (Visual)       &  2.7 &  0.9 &  6.4 &  2.6 &   3.3 &  0.1 \\
BYOC-Geo            & 4.8 &  2.3 & 10.6 &  5.4 &  7.2 &  0.5 \\
BYOC                & 3.8 &  1.7 &  8.7 &  4.3 &  5.6 &  0.3 \\
\bottomrule
\end{tabular}
}
\label{tab:analysisfake}
\caption{
\textbf{Random Visual Features are surpsingly good for registration.}
We achieve good alignment with random visual features, but not geometric features. This pattern holds after training: it is better to align using visual features. 
}
\label{tab:analysis}

\end{table}
}

\lsparagraph{What is the impact of the transformation estimator?}
While we observe that RANSAC improves the performance of FPFH and FCGF compared to the Weighted Procrustes, we see the opposite pattern with our approach. 
This is due to the fact that our model is trained specifically on a registration loss on filtered correspondence. As a result, Lowe's ratio becomes a very effective method of filtering our correspondences while being less effective for other approaches. 

\lsparagraph{How good are random features?}
We find that random visual features can serve as a strong baseline for point cloud registration on ScanNet, as shown in Fig~\ref{fig:intial_corr} and Table~\ref{tab:analysis}. 
This is suprising since random visual features perform on-par with FCGF. This explains why our method is capable of achieving this performance without any supervision. 
We also find that after training, our visual features achieve the highest registration performance. 
Those results point that visual features are better descriptors for registration, but it is unclear if this a fundemntal advantage or if the performance gap can be be resolved through better architectures or training schemes for geometric feature learning.

\subsection{Correspondence Estimation} 
\label{sec:exp_corr}

We now examine the quality of the correspondences estimated by our method.  
We evaluate our approach on the 3D Match geometric registration benchmark and follow the evaluation protocol proposed by Deng~\etal~\cite{deng2018ppfnet} of evaluating the correspondence recall.
Intuitively, feature-match recall measures the percentage of point cloud pairs that would be registered accurately using a RANSAC estimator by guaranteeing a minimum percentage of inliers. 

\lsparagraph{Baselines. } 
We compare our approach against three sets of baselines. 
The first set are hand-crafted features based on the local geometry around each point~\cite{rusu2009fast,salti2014shot,tombari2010usc}. 
The second set are supervised approaches that use known pose to sample ground-truth correspondences and apply a metric learning loss to learn features for geometric registration. 
Finally, the third set are unsupervised approaches trained on reconstructed scenes.
While those approaches do not directly use ground-truth pose during training, their training data (reconstructed scenes) is generated by aligning 50 depth maps into a single point cloud. Hence, while those approaches do not use pose supervision explicity, pose information is \textit{needed} to generate their data. We refer to those approaches as \textit{scene-supervised}.

\begin{table}[t]
\centering
\resizebox{\linewidth}{!}{

    \begin{tabular}{l | c c | c cc }
    \toprule
     & \multicolumn{2}{c}{Training Data} &  \multicolumn{2}{c}{FMR}  
    \\

    & Dataset & Data Format & Recall & St. Dev.  
    \\    \midrule

    SHOT~\cite{salti2014shot}           & - & - & 0.238 & 0.109 \\
    USC~\cite{tombari2010usc}           & - & - & 0.400 & 0.125 \\
    
    FPFH~\cite{rusu2009fast}            & - & - & 0.481 & 0.150 \\
    FPFH~\cite{rusu2009fast} (corr)     & - & - & 0.462 & 0.198 \\

    \midrule
    3D Match~\cite{zeng20163dmatch}             & 3D Match & Depth + Pose  & 0.596 & 0.088 \\
    PPFNet~\cite{deng2018ppfnet}                & 3D Match & Depth + Pose  & 0.623 & 0.108  \\
    PerfectMatch~\cite{gojcic2019perfect}       & 3D Match & Depth + Pose  & 0.947 & 0.027 \\
    FCGF~\cite{choy2019fully}                             & 3D Match & Depth + Pose  & 0.952 & 0.066 \\
    FCGF~\cite{choy2019fully} (corr)                      & 3D Match & Depth + Pose  & 0.932 & 0.104  \\
    \midrule

    % Unsupervised on reconstructed scenes
    CGF~\cite{ranftl2018deepfundamental}        & SceneNN  & Scenes & 0.582 & 0.142    \\
    PPF-FoldNet~\cite{deng2018ppffoldnet}       & 3D Match & Scenes & 0.718 & 0.105   \\
    3D PointCapsNet~\cite{zhao20193dpointcaps}  & 3D Match & Scenes & 0.787 & 0.062   \\ 

    \midrule
    BYOC w/ no filtering    & ScanNet   & RGB-D  & 0.662 & 0.225    \\
    BYOC                    & 3D Match  & RGB-D  & 0.690 & 0.172    \\
    BYOC-Geo                & ScanNet   & Depth  & 0.786 & 0.195    \\
    BYOC                    & ScanNet   & RGB-D  & 0.766 & 0.181    \\
    BYOC-Rot                & ScanNet   & RGB-D  & 0.827 & 0.150    \\
    \bottomrule
    \end{tabular}
}
\label{tab:fmr_3dmatch}
\caption{\textbf{Feature-Match Recall on 3D Match.}
Our approach achieves better recall than hand-crafted and scene-supervised approaches while being competitive with supervised approaches.
}

\end{table}

\lsparagraph{Evaluation Metrics.}
Given a set of correspondences $\mathcal{C}$, $FM(\mathcal{C})$ evaluates whether the percentage of inliers exceeds $\tau_2$, where an inlier correspondence is defined as having a residual error less than $\tau_1$ given the ground-truth transformation $\mathbf{T}^{*}$. Feature-match recall is the percentage of point cloud pairs that have a successful feature matching. 
\begin{equation}
FM(\mathcal{C}) = 
% \frac{1}{M} \sum\limits_{s=1}^M  \Bigg( 
\Big[\frac{1}{|\mathbf{\mathcal{C}|}} 
\sum\limits_{(p, q) \in \mathbf{\mathcal{C}}} 
\mathbf{\mathbbm{1}}
\big( ||\mathbf{x}_p{-}\mathbf{T}^{*}\mathbf{x}_q|| < \tau_1\big)\Big] > \tau_2 
% \Bigg)
\end{equation}
Similar to \cite{choy2019fully,deng2018ppffoldnet,deng2018ppfnet}, we calculate feature-match recall over all view pairs using $\tau_{1}=10$ cm and $\tau_2 = 5\%$.
Prior approaches often generate feature sets without any specified means of filtering them. As a result, they define the correspondence set as the set of all nearest neighbors.
Unlike prior work, our approach outputs a small set of correspondences after ranking them using Lowe's ratio test.

\begin{figure*}[t]
\begin{center}
  \includegraphics[width=0.92\linewidth]{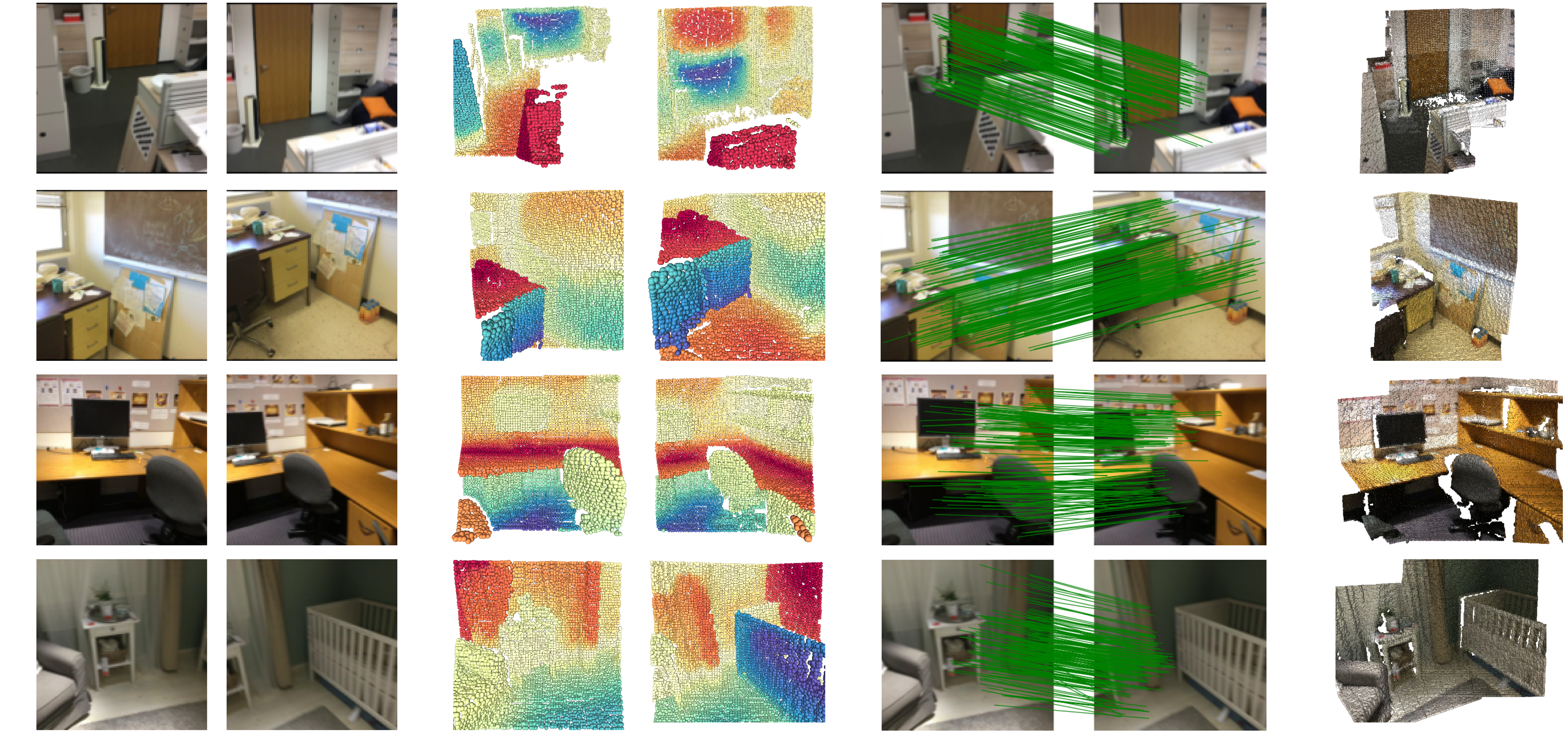}
\end{center}
    \caption{\textbf{{BYOC}'s geometric features allow for accurate registration by mapping corresponding points to similar feature vectors.} Our approach learns informative geometric features of the scene. We visualize our features by mapping them to colors using t-SNE~\cite{van2008visualizing}. We find that the learned features appear to delineate objects such as chairs and floor edges. This results in the accurate registrations shown in the last column. Our approach takes uncolored point clouds as input; images and color point cloud are presented to aid visualization. }
\label{fig:qualitative}
\end{figure*}

\lsparagraph{Results. }
We find that our approach achieves a high feature-match recall, outperforming traditional approaches and scene-supervised approaches, while being competitive with supervised approaches. 
It is worth emphasizing that we achieve this performance while training on the raw RGB-D or depth scans without requiring any additional annotation or post-processing of the data. 
We achieve the best performance by training on ScanNet with rotation augmentations. Rotation augmentation gives us a small boost, despite resulting in lower registration performance on ScanNet. 
This can be explained by the differing data distribution between the two dataset: 3D Match benchmark has larger transformations than those observed between video frames. Hence, data augmentation becomes very useful. 

We also observe the interesting pattern that the model trained on only geometric correspondence generalizes better to 3D Match despite doing worse on ScanNet. 
One explanation for this discrepency is that bootstrapping with visual correspondences biases the model towards representing features that are meaningful in both modalities.
Such representations might be more dataset specific, hindering across-dataset generalization.
This finding also opens up the possibility of scaling datasets that only include depth video; \eg, lidar datasets.

While our best configuration outperforms all the scene-supervised approaches, 
we achieve performance that is competitve with the scene-supervised approaches when we evaluate all our features (\textit{no filtering}).
We observe that when we attempt to filter the correspondences for FPFH or FCGF, their performance deteriorates.
This is consistent with some of the reported results by~\cite{deng2018ppffoldnet} where using a larger number of features improved their performance. 
Hence, it is unclear how correspondence filtering  would affect their performance. 
Due to the lack of an official implementation of PPF-FoldNet and the complexity of their approach, we were unable to run additional experiments to better understand the impact of the training data and correspondence filtering on the learning process. This affects both PPF-FoldNet and 3D Point Capsule Networks since the latter approach replaces the encoder in PPF-FoldNet.

\section{Conclusion}
\label{sec:conclusion}
We propose BYOC: a self-supervised approach to point cloud registration. 
Our key insight is that randomly initialized CNNs provide us with features that are good enough to bootstrap visual and geometric feature learning through point cloud registration. 
Our approach takes advantage of pseudo-correspondence labels that are obtained from a visual encoder to apply feature similarity loss to both the geometric and visual encoders. 
At test time, we only use the geometric encoder to register point clouds without relying on color or image information. 

Our approach is both simple and fast: we rely on a fast sparse 3D convolutional encoder to extract features, use a ratio test to estimate correspondence, and align the features using SVD. 
This deviates from current state-of-the-art approaches that use expensive prepossessing techniques~\cite{deng2018ppfnet,deng2018ppffoldnet,zhao20193dpointcaps}, learn separate networks for correspondence estimation~\cite{choy2020deep,gojcic2020learning,ranftl2018deepfundamental}, and use RANSAC as the transformation estimation. Furthermore, we only depth or RGB-D videos to train our model. This allows us to train on any dataset of such format, not only ones that can be accurately registered by traditional SfM pipelines.

\vspace{6pt} \noindent
\textbf{Acknowledgments}
We would like to thank Richard Higgins and Karan Desai for many helpful discussions and feedback on early drafts of this work.

{\small
\bibliographystyle{ieee_fullname}
\bibliography{bibliography}
}

\end{document}